\documentclass[10pt,twocolumn,letterpaper]{article}
\usepackage{times}
\usepackage{graphicx} 
\usepackage{subfigure} 
\usepackage{amsmath}
\usepackage{epstopdf}
\usepackage{float}
\usepackage{courier}
\usepackage{cite}

\usepackage{dblfloatfix}

\usepackage[utf8]{inputenc}
\usepackage[english]{babel}
 
\usepackage{algorithm}
\usepackage{algorithmic}

\usepackage{hyperref}

\usepackage{iccv}
\iccvfinalcopy
\usepackage[section]{placeins}


\usepackage{setspace}

\begin{document}

\title{\textbf{Abnormality Detection and Localization in Chest X-Rays\\using Deep Convolutional Neural Networks}}

\author{Mohammad Tariqul Islam$^1$, Md Abdul Aowal$^1$, Ahmed Tahseen Minhaz$^1$, Khalid Ashraf$^2$\\
$^1$Semion, House 167, Road 3, Mohakhali DOHS, Dhaka, Bangladesh.\\
$^2$Semion, 1811 Francisco St., St 2,  Berkeley, CA 94703, USA.\\
{\tt\small \{mhdtariqul, aowal.eee, tahseenminhaz92\}@gmail.com, \{khalid\}@semion.ai }
}

\maketitle





\begin{abstract}

Chest X-Rays (CXRs) are widely used for diagnosing abnormalities in the heart and lung area. Automatically detecting these abnormalities with high accuracy could greatly enhance real world diagnosis processes. Lack of standard publicly available dataset and benchmark studies, however, makes it difficult to compare and establish the best detection methods. In order to overcome these difficulties, we have used the publicly available Indiana chest X-Ray dataset, JSRT dataset and Shenzhen Dataset and studied the performance of known deep convolutional network (DCN) architectures on different abnormalities. We employed heat maps obtained from occlusion sensitivity as a measure of localization in the CXRs. We find that the same DCN architecture doesn't perform well across all abnormalities. Shallow features or earlier layers consistently provide higher detection accuracy compared to deep features. We have also found ensemble models to improve classification significantly compared to single model. Combining these insight, we report the highest accuracy on chest X-Ray abnormality detection on this dataset. We find that in the cardiomegaly classification task, where comparison could be made, the deep learning method improves the accuracy by a staggering 17 percentage point compared to rule based methods. We applied the techniques developed along the way to the problem of tuberculosis detection on a different dataset and achieved the highest accuracy on that task. Our localization experiments using these trained classifiers show that for spatially spread out abnormalities like cardiomegaly and pulmonary edema, the network can localize the abnormalities successfully most of the time. One remarkable result of the cardiomegaly localization is that the heart and its surrounding region is most responsible for cardiomegaly detection, in contrast to the rule based models where the ratio of heart and lung area is used as the measure. We believe that through deep learning based classification and localization, we will discover many more interesting features in medical image diagnosis that are not considered traditionally. 

\end{abstract}


\section{Introduction}
\label{intro}

Medical X-rays are one of the first choices for diagnosis due to its ``ability of revealing some unsuspected pathologic alterations, its non-invasive characteristics, radiation dose and economic considerations" \cite{campadelli2005lung}. X-Rays are mostly used as a preliminary diagnosis tool. There are many benefits of developing computer aided detection (CAD) tools for X-Ray analysis. First of all, CAD tools help the radiologist to make a quantitative and well informed decision. As the data volume increases, it will become increasingly difficult for the radiologists to go through all the X-Rays that are taken maintaining the same level of efficiency. Automation and augmentation is severely needed to help radiologists maintain the quality of diagnosis. 

Over the past decade, a number of research groups have focused on developing CAD tools to extract useful information from X-Rays. Historically, these CAD tools depended on rule based methods to extract useful features and draw inference based on them. The features are often useful for the doctor to gain quantitative insight about an X-Ray, while inference helps them to connect those abnormal features to certain disease diagnosis. However, the accuracy of these CAD tools has not achieved a significantly high level to work as independent inference tool. Thus CAD tools in X-Ray analysis are left as mostly providing easy visualization functionality.

In recent time, deep learning has achieved superhuman performance on a number of image based classification\cite{krizhevsky2012imagenet,he2016deep}. This success in recognizing objects in natural images has spurred a renewed interest in applying deep learning to medical images as well. A number of reports recently have emerged where indeed superhuman accuracies were obtained in a number of abnormality detection tasks. This success of classifying abnormalities in images have not translated to other radiological modalities mainly because of the absence of large standard datasets. Creation of high quality and orders of magnitude larger dataset will certainly drive the field forward. 

In this work, we report DCN based classification and localization on the publicly available datasets for chest X-Rays. Our contributions are the following: 
\begin{itemize}
\item We show a 17 percentage point improvement in accuracy over rule based methods for Cardiomegaly detection using ensemble of deep convolutional networks.
\item Multiple random train/test data split achieve robust accuracy results when the number of training examples are low.
\item Shallow features or earlier layers perform better than deep features for classification accuracy. 
\item Ensemble of DCN models performs better than single models. However, mix of rule based and DCN ensemble model degraded accuracy.
\item Sensitivity based localization provides correct localization for spatially spread out diseases.
\item Results of 20 different abnormalities which we believe will serve as a benchmark for other studies to be compared against.
\item Direct application of the methods developed in the paper on the Shenzen dataset achieve the highest accuracy  for tuberculosis detection. 
\end{itemize}   

The paper is organized as follows. In section~\ref{relWork} we overview of the related work. In section~\ref{experiments}, we describe the dataset, analysis method, evaluation figure of merits and the localization method used. Then in section~\ref{results_classification}, we present our results on single and ensemble models and critique various issues discussed above. In section~\ref{results_localization}, we describe the localization results and discuss their performance. The cardiomegaly detection and tuberculosis detection are discussed in detail along with comparison sections~\ref{card_describe} and~\ref{tb_describe}. Finally we conclude summarizing our results in section~\ref{conclusion}. The conclusions in the paper are derived by analyzing two representative abnormalities i.e. cardiomegaly and pulmonary edema. The classification results for other abnormalities are given in the supplementary materials.   


\section{Related Works}
\label{relWork}
Local binary pattern (LBP) features were employed in segmented images to classify normal vs. pathology on CXRs in \cite{carrillo2016computer} for early detection purposes. The dataset used in the study was private and contained 48 images total. 
In \cite{candemir2016automatic}, image registration technique was used to localize the heart and lung region and then computed radiographic index like cardiothoracic ratio (CTR), cardiothoracic area ratio (CTAR) to classify cardiomegaly from the X-ray images. In \cite{candemir2014lungSegmentation} lung segmentation was performed using 247 images from JSRT, 138 images from Montgomery and 397 images from the India dataset with segmentation accuracies of 95.4\%, 94.1\%, and 91.7\% respectively.
Jaeger \emph{et. al}~\cite{jaeger2014automatic} segmented lungs using graph cut method and used large features sets both from the domain of object detection and content based image retrieval for early screening of tuberculosis (TB) and made the databases public. Additionally, few other works on TB screening has been conducted using the public datasets~\cite{lopes2017pre} and using additional data along with the public datasets~\cite{hwang2016novel,lakhani2017deep}. They achieved near human performance in detecting TB.  Gabor filter features were extracted from histogram equalized CXRs in \cite{kumar2014distinguishing} in order to detect pulmonary edema using 40 pulmonary edema and 40 normal images and achieved 97\% accuracy. The dataset is private hence the accuracy cannot be compared. In an attempt to identify multiple pathologies in a single CXR, bag of visual words is constructed from local features which are fed to probabilistic latent semantic analysis (PLSA) pipeline \cite{zare2013automatic}. They used the ImageClef dataset and clustered various types of X-Rays present in the dataset. However, they didn't detect any abnormality in the paper. 
In a view to classifying abnormalities in the CXRs, a cascade of convolutional neural network (CNN) and recurrent neural network (RNN) are employed \cite{shin2016learning} on the Indiana dataset chest X-Rays. However, no test accuracy was given nor any comparison with previous results was discussed. Hence it was impossible to determine the robustness of the results. 
Usage of pre-trained \textit{Decaf} model in a binary classifier scheme of normal vs. pathology, cardiomegaly, mediastinum and right pleural effusion have been attempted \cite{bar2015chest}. This work was reported on a private dataset, and hence no comparison can be made. 

\subsection{Deep Learning on Medical Image Analysis}
\label{dlMedImg}
A detailed survey of deep learning in medical image analysis can be found in \cite{litjens2017survey}. Localization of cancer cells is demonstrated in \cite{wang2016deep}. Using inception network, human level diabetic retinopathy detection is shown in \cite{gulshan2016development}. Using a multiclass approach, inception network is used in \cite{esteva2017dermatologist}, to obtain human level skin cancer detection.

\section{Experiments}
\label{experiments}
\subsection{Datasets}
The three publicly available datasets for our studies in this paper are:
\begin{itemize}
\item Indiana Dataset~\cite{demner2016preparing}: Set consists of 7284 CXRs, both frontal and lateral images with disease annotations, such as cardiomegaly, pulmonary edema, opacity or pleural effusion. Indiana Set is collected from various hospitals affiliated with the Indiana University School of Medicine. The set is publicly available through Open-i SM, which is a multimodal (image + text) biomedical literature search engine developed by U.S. National Library of Medicine. A typical example of a normal CXR (left) and a CXR with cardiomegaly abnormality (right) is shown in Fig. \ref{fig:indiana_example}. Visually, it can be observed that the heart in the cardiomegaly example is quite big compared to that of the normal CXR.

\begin{figure}[t]
\centering
\begin{minipage}[t!]{0.2\textwidth}
    \includegraphics[width=\textwidth]{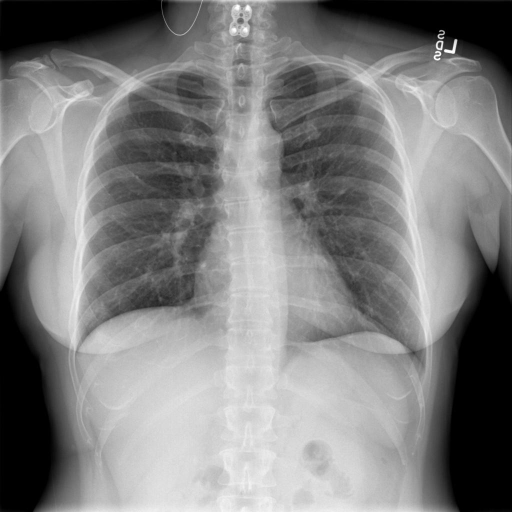}
    \end{minipage}
\begin{minipage}[t!]{0.2\textwidth}
    \includegraphics[width=\textwidth]{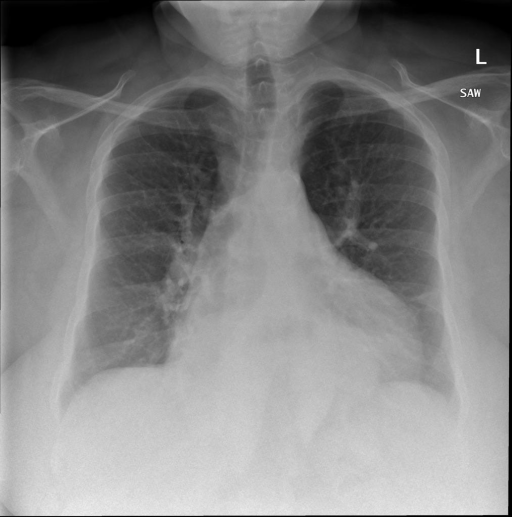}
    \end{minipage}
    \vspace{0.1in}
  \caption{An example of Normal CXR (left) and an example of a cardiomegaly CXR (right) from Indiana dataset. The pathology in the right CXR can be easily distinguished from the abnormal size and shape of the heart.}
  \label{fig:indiana_example}
\end{figure}

\item JSRT Dataset~\cite{shiraishi2000development,van2006segmentation}: Set compiled by the Japanese Society of Radiological Technology (JSRT). The set contains 247 chest X-rays, among which 154 have lung nodules (100 malignant cases, 54 benign cases), and 93 have no nodules. All X-ray images have a size of $2048\times2048$ pixels and a gray-scale color depth of 12 bit. The pixel spacing in vertical and horizontal directions is 0.175 mm. The JSRT set is publicly available and has gold standard masks~\cite{van2006segmentation} for performance evaluation.

\item Shenzhen Dataset~\cite{jaeger2014two}: This set is compiled at Shenzhen No.3 People’s Hospital, Guangdong Medical College, Shenzhen, China. The recorded frontal CXRs are classified into two categories: normal and tuberculosis (TB). In a one month period, 326 normal cases and 336 cases with tuberculosis have been recorded from the outpatient clinics comprising a total of 662 CXRs in the dataset. The clinical reading of each of the CXRs is also provided.

\end{itemize}

\begin{figure*}[h]
\centering
\includegraphics[scale=0.27]{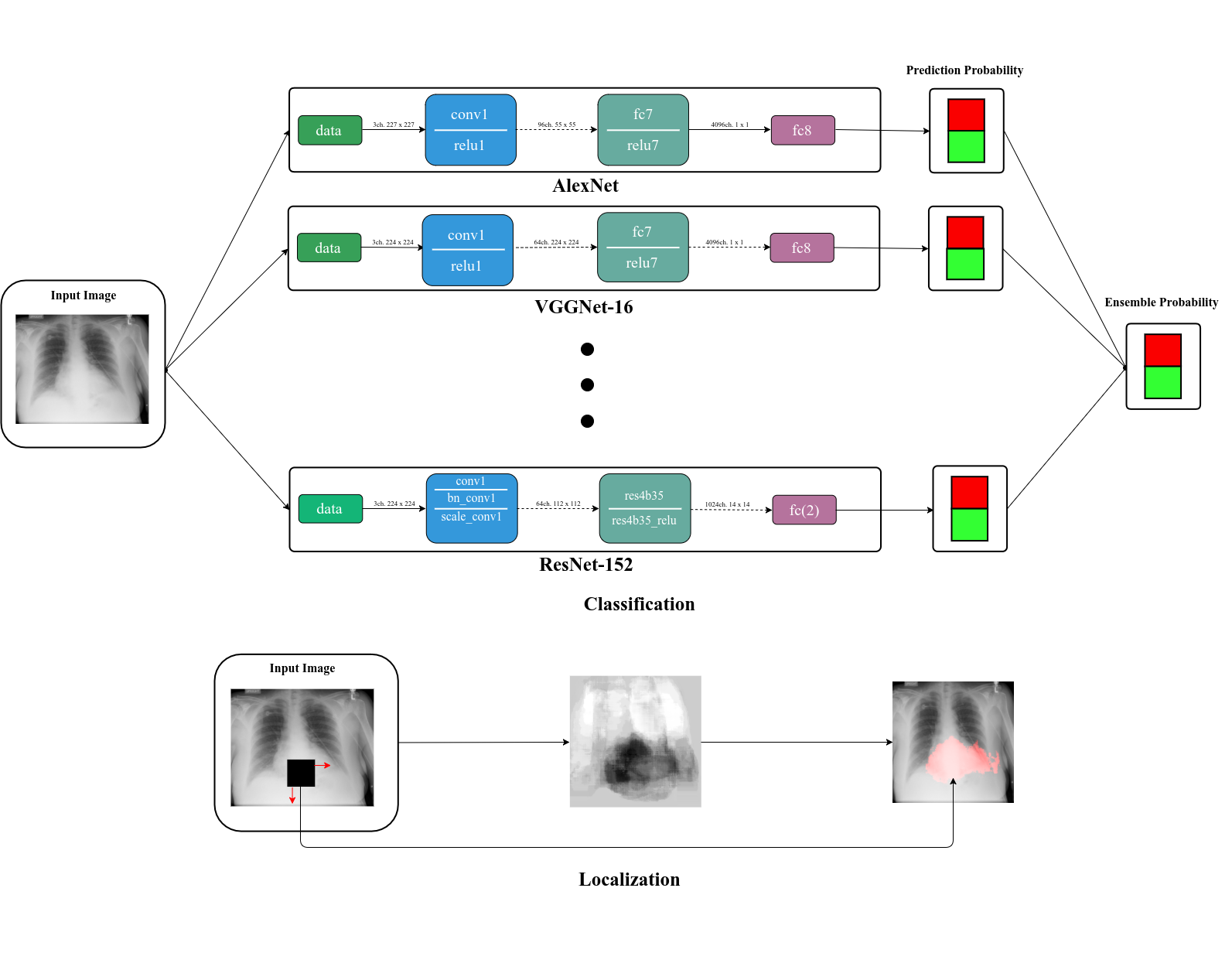}\\
\caption{Summary of the proposed classification and localization method. In the classification stage ensemble is obtained using different DCN models. The localization is obtained by exploiting occlusion sensitivity of the classifiers.} \label{fig:dnn_architecture}
\end{figure*}

\subsection{Deep Convolution Network Models}
As described in section \ref{dlMedImg}, deep convolutional networks (DCN) have achieved significantly higher accuracy than previous methods in disease detection in various diagnostic modalities. In many cases, these accuracies have surpassed human detection capabilities. Here, we explore the performance of various DCNs for heart disease detection on chest X-Rays. We use binary classification of Cardiomegaly and Pulmonary Atelectasis against normal chest X-Rays as representative examples. Results for other diseases are given in the supplementary materials. We explored several DCN models, e.g, AlexNet \cite{krizhevsky2012imagenet}, VGG-Net \cite{Simonyan14c} and ResNet \cite{he2016deep}. These models vary in the number of convolution layers used and achieve higher classification accuracy as the number of convolution layers is increased. Specifically, ResNet and its variants have achieved superhuman performance on the celebrated ImageNet dataset. In the experiments we have extracted features from one of the layers of the DCN. We have frozen all the layers upto this layer and added a binary classifier layer to detect the abnormality.
The second fully connected layer has been selected for feature extraction in AlexNet, VGG-16 and VGG-19 networks.
The features from the ResNet-50, ResNet-101 and ResNet-152 are extracted from the \texttt{res4f}, \texttt{res4b22} and \texttt{res4b35} layers respectively. All the DCN models have been implemented in Tensorflow and have been finetuned using Adam optimizer~\cite{kingma2014adam} with learning rate $0.001$. The weights of the networks AlexNet, and VGG were obtained from the respective project pages, while weights of the ResNet models were obtained from MatConvNet Pre-train Library~\footnote{http://www.vlfeat.org/matconvnet/pretrained/}.

\subsection{Evaluation Metrics}

The quality of detection was evaluated in terms for four measures: accuracy, area under receiver operating characteristics (ROC) curve (AUC), sensitivity and specificity. The accuracy is the ratio of number of correctly classified samples to total samples. Unless otherwise stated, classifier threshold is set to $0.50$ in the reported values of accuracy, sensitivity and specificity. ROC curve is the graphical plot of true positive rate (TPR) vs false positive rate (FPR) of a binary classifier when classifier threshold is varied from $0$ to $1$. The number of pathological samples that are correctly identified as pathological sample by the classifier is called true positive (TP). The number of pathological samples that are incorrectly classified as normal by the classifier is called false negative (FN). The number of normal samples that are correctly classified as normal is called true negative (TN), and in a similar fashion, the number of normal samples that are incorrectly identified as pathological samples is called false positive (FP). True positive rate (TPR) is the proportion of pathological samples that are correctly identified as pathological sample, given as
\begin{align}
\textnormal{TPR}=\textnormal{sensitivity}=\frac{\textnormal{TP}}{\textnormal{TP}+\textnormal{FN}}
\end{align}
TPR is also called sensitivity which is called such as this measure shows the degree to which does not miss a pathological sample. False positive rate (FPR) is proportion of normal samples that are incorrectly identified as pathological samples, given as,
\begin{align}
\textnormal{FPR}=1-\textnormal{specificity}=\frac{\textnormal{FP}}{\textnormal{FP}+\textnormal{TN}}
\end{align}
The measure specificity shows the degree to which the classifier correctly identifies normal samples as normal. The objective of a classifier to attain high sensitivity as well as specificity so that the classifier attains low diagnosis error.

\subsection{Localization Scheme} \label{sec:loc_scheme}

The sensitivity of softmax score to occlusion of a certain region in the chest X-Ray was used to find which region in the image is responsible for the classification decision. We followed the localization using occlusion sensitivity described in~\cite{zeiler2014visualizing}. In this experiment, a patch of square size is occluded in the CXRs and is observed whether the classifier can detect pathology in the presence of the occlusion. If the region corresponding pathology is occluded then the classifier should no longer detect the pathology with higher probability and thus this drop in probability indicates that the pathology is located at the location of the occlusion. This occluded region is slid through the whole CXR and thus a probability map of the pathology corresponding to the CXR is obtained. The regions where the probabilities are below a certain threshold indicates that the pathology is likely to be occupying that region. Thus, the pathology in the CXR can be localized.

The overall classification scheme and localization scheme is visualized in Fig.~\ref{fig:dnn_architecture}. In summary, the classification scheme (top) is ensemble of different types of DCNS and the localization (bottom) is obtained from the overlapping occlusions.

\section{Results}
\label{results}

\begin{table*}[t]
\caption{Accuracy, AUC, sensitivity and specificity using standard DCNs on Cardiomegaly abnormality.} 
\label{ShallowvDeep}
\vskip 0.15in
\begin{center}
\begin{small}
\begin{sc}
\begin{tabular}{lcccccr}
\hline
 & Accuracy (\%) & AUC & Sensitivity (\%) & Specificity (\%) \\
\hline
Alex Net    & $86.00\%$ 	& $0.92$  	& $86.00\%$ 	& $86.00\%$ \\
VGG-16      & $86.00\%$ 	& $0.87$  	& $\bf96.00\%$ & $76.00\%$ \\
VGG-19      & $\bf92.00\%$ & $\bf0.94$  	& $92.00\%$ 	& $92.00\%$ \\
ResNet-50   & $87.00\%$ 	& $0.93$  	& $94.00\%$ 	& $80.00\%$\\
ResNet-101  & $\bf92.00\%$ & $0.92$  	& $88.00\%$ 	& $\bf96.00\%$\\
ResNet-152  & $90.00\%$ 	& $0.91$  	& $92.00\%$ 	& $88.00\%$\\

\hline
\end{tabular}
\end{sc}
\end{small}
\end{center}
\vskip -0.1in
\end{table*}

\begin{table*}[t]
\caption{Accuracy, AUC, sensitivity and specificity using standard DCNs using dropout on features on Cardiomegaly Abnormality.}
\label{ShallowvDeep_drop}
\vskip 0.15in
\begin{center}
\begin{small}
\begin{sc}
\begin{tabular}{lccccr}
\hline
 & Accuracy (\%) & AUC & Sensitivity (\%) & Specificity (\%) \\
\hline
Alex Net    & $88.00\%$ 	& $\bf0.94$  & $88.00\%$ 	& $88.00\%$ \\
VGG-16      & $\bf89.00\%$ & $0.90$  	& $90.00\%$ 	& $88.00\%$ \\
VGG-19      & $88.00\%$ 	& $0.88$  	& $86.00\%$ 	& $90.00\%$ \\
ResNet-50   & $88.00\%$ 	& $0.90$  	& $88.00\%$ 	& $88.00\%$\\
ResNet-101  & $87.00\%$ 	& $0.91$  	& $82.00\%$ 	& $\bf92.00\%$\\
ResNet-152  & $87.00\%$ 	& $0.88$  	& $\bf92.00\%$ & $82.00\%$\\

\hline
\end{tabular}
\end{sc}
\end{small}
\end{center}
\vskip -0.1in
\end{table*}

\subsection{Classification}
\label{results_classification}
\subsubsection{Classification using single models}

Our first experiment use single model with DCNs fine-tuned from a model trained on ImageNet. Detection of cardiomegaly is done only for the frontal CXR images from the Indiana Dataset. It contains 332 frontal CXRs with cardiomegaly. In order to balance the binary classification, 332 normal frontal CXRs have been selected randomly from the database. Of these images, 282 of each class have been selected for training and 50 of each class for testing. 

In addition to training the DCNs, we also performed rule based features for cardiomegaly detection. Overall, we ran experiments with the following characteristics: (1) The NNs are fine-tuned on the Indiana dataset, (2) The NNs are fine-tuned using dropout technique \cite{srivastava2014dropout}, (3) The fusion of NN feature and rule based features, and (4) The fusion of NN feature and rule based feature trained using dropout technique. The results are summarized in tables \ref{ShallowvDeep}-\ref{ShallowvDeep_drop}. 

In table \ref{ShallowvDeep}, the results obtained by fine-tuning the DCNs are shown. We find that deeper models like VGG-19 and ResNet improve the classification accuracy significantly. For example, the accuracy of Cardiomegaly detection improves by 6 percentage point from that using AlexNet when VGG-16 and ResNet-101 are used. In order to understand the robustness of these  results, we further calculate the sensitivity, specificity, sensitivity vs 1-specificity curve and derive the area under curve (AUC) metric for classification using different networks. We find that although ResNet-101 gives the highest specificity and VGG-16 gives the highest sensitivity, VGG-19 gives an overall better performance with the highest AUC of $0.94$. The AUC calculated using VGG-19 is at least one percentage point higher than the other networks considered here.

Adding dropout improves the classification accuracy of the shallower networks but degrades the performance of deep models. We find that VGG-16 and AlexNet achieve the highest accuracy and AUC respectively when dropout is used as shown in table \ref{ShallowvDeep_drop}. On the contrary, the accuracy of deeper models like ResNet-101 and VGG-19 drops by about 4 percentage points.

\begin{figure}[t]
\centering
\includegraphics[width=8cm, height=6cm]{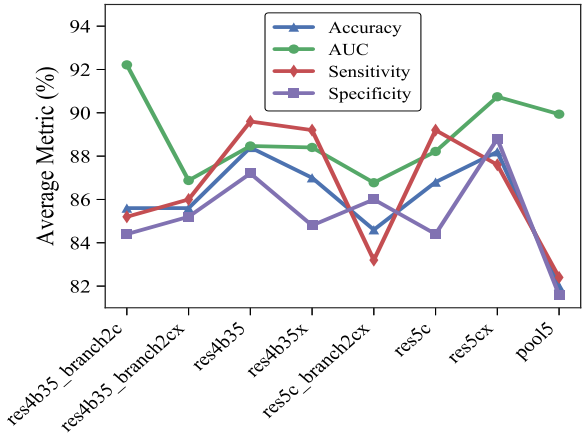}\\
\caption{Performance for features extracted from different layers of ResNet-152 for Cardiomegaly detection. Features from earlier layers provides much better accuracy than the final pooling layer (\texttt{pool5}) with highest from \texttt{resb35} features.} 
\label{fig:ResNet_layers}
\end{figure}

For all these experiments, we found that taking features from earlier layers compared to later layers improve accuracy by 2 to 4 percentage points. Shallow DCN features are often useful for detecting small objects in images \cite{ashraf2017ShallowNN}. Our findings are similar for chest X-Ray abnormality classification as well. As an example, we are showing the performance obtained by taking features from different layers of ResNet-152 model. The candidate layers are chosen from the 4th, 5th and final stage of the network based on what type of operations they perform. The chosen layers and their corresponding operations are listed in Table~\ref{tab:ResNet_table}. The notation of the layers is based on the pre-trained model obtained from MatConvNet Pre-train Library. We trained five models to detect cardiomegaly using features from each of the layers and the average performance of these features in terms of accuracy, AUC, sensitivity, and specificity for Cardiomegaly detection are shown in Fig.~\ref{fig:ResNet_layers}. It can be observed that the performance of the final pooling layer (\texttt{pool5}) is degraded compared to the other layers in terms of accuracy, sensitivity and specificity. In particular features from residual connections (\texttt{res4b35}, \texttt{res5c}) and ReLU (\texttt{res4b35x}, \texttt{res5cx}) are considerably better with features from \texttt{res4b35} providing highest accuracy. Similar observations are made for other ResNet variants, VGG nets and AlexNet.

\begin{table}[t]
\centering
\caption{Candidate layers and their operation types chosen from ResNet-152 to test the features obtained from these layers.}
\label{tab:ResNet_table}
\vskip 0.15in
\begin{center}
\begin{small}
\begin{sc}
\begin{tabular}{lccr}
\hline
Layer Name & Stage & Operation  \\
\hline
\texttt{res4b35\_branch2c}	& 4th   & Convolution \\
\texttt{res4b35\_branch2cx} & 4th   & Batch Normalization \\
\texttt{res4b35}      		& 4th 	& Residual Connection \\
\texttt{res4b35x} 			& 4th   & ReLU \\
\texttt{res5c\_branch2c}  	& 5th	& Convolution \\
\texttt{res5c\_branch2cx}	& 5th	& Batch Normalization \\
\texttt{res5c}				& 5th	& Residual Connection \\
\texttt{res5cx}				& 5th	& ReLU \\
\texttt{pool5}				& Final	& Average Polling \\

\hline
\end{tabular}
\end{sc}
\end{small}
\end{center}
\vskip -0.1in
\end{table}


In addition to the DCN features, we experimented with DCN and rule based feature fusion for single model classification. The rule based features that were used in the study are 1D-cardio-thoracic ratio (CTR), 2D-cardio-thoracic ratio and cardio-thoracic area ratio (CTAR) \cite{candemir2016automatic}. 

1D-CTR is the ratio between the maximum transverse cardiac diameter and the maximum thoracic diameter measured between the inner margins of ribs, which is formulated as,
\begin{align}
\label{1D-CTR}
\textnormal{1D-CTR} = \frac{\textnormal{Maximum transverse cardiac diameter}}{\textnormal{Maximum thoracic diameter}}
\end{align}

The 2D-CTR is the ratio between the perimeter of the heart region to the perimeter of the entire thoracic region and formulated as
\begin{align}
\label{2D-CTR}
\textnormal{2D-CTR} = \frac{\textnormal{Perimeter of Heart}}{\textnormal{Perimeter of Thoracic Region}}
\end{align}

while CTAR, the ratio between area of the heart region to the sum of the area of the left and right lung region, is formulated as
\begin{align}
\label{CTAR}
\textnormal{CTAR}=\frac{\textnormal{Area of Heart}}{\textnormal{Area of Left Lung}+\textnormal{Area of Right Lung}}.
\end{align}

In the experiments involving rule based features, we concatenated the features with the features extracted from a DCN and trained a fully connected layer to detect cardiomegaly. However, the results degraded and hence are not shown here.

Observation from these single model classification results is that different figure of merit is maximized by different DCNs. We wanted to explore if this is expected or due to some limitation of the data or training process itself. Hence, rather than taking a single train-test split of the data, we randomly split the train-test data and trained nine different model for each architecture. Then we calculated the mean and standard deviation for the figure of merits of interest. The results can be seen in table~\ref{Cardiomegaly}. We find that after averaging the nine random train-test sample results, a clear trend emerges where a single model, ResNet-152 in this case, achieves the highest accuracy, AUC and sensitivity. The mean specificity for ResNet-152, in this case is close to the highest number, however, the max specificity is indeed highest for ResNet-152.    

\begin{table*}[t]
\caption{Mean and standard deviation of accuracy, AUC, sensitivity and specificity for different train-test split using standard DCNs on Cardiomegaly. Deeper network achieves consistently better accuracy and AUC.}

\label{Cardiomegaly}
\vskip 0.15in
\begin{center}
\begin{small}
\begin{sc}
\begin{tabular}{lcccccr}
\hline
 & Accuracy (\%) & AUC & Sensitivity (\%) & Specificity (\%) \\
\hline
Alex Net    & $84.73 \pm 3.03\%$ & $0.91 \pm 0.03$  & $85.53 \pm 4.75\%$ & $83.93 \pm 3.15\%$ \\
VGG-16      & $87.37 \pm 1.76\%$ & $0.91 \pm 0.02$  & $87.20 \pm 3.00\%$ & $87.53 \pm 2.07\%$ \\
VGG-19      & $87.97 \pm 2.04\%$ & $0.91 \pm 0.02$  & $88.60 \pm 2.26\%$ & $87.33 \pm 3.72\%$ \\
ResNet-50   & $87.33 \pm 1.96\%$ & $0.91 \pm 0.02$  & $86.87 \pm 3.76\%$ & $\bf 87.80 \pm 3.75\%$ \\
ResNet-101  & $86.30 \pm 2.00\%$ & $0.91 \pm 0.02$  & $88.60 \pm 2.26\%$ & $87.33 \pm 3.72\%$ \\
ResNet-152  & $\bf 88.03 \pm 2.27\%$ & $\bf 0.92 \pm 0.02$  & $\bf 88.87 \pm 5.77\%$ & $87.20 \pm 4.55\%$ \\

\hline
\end{tabular}
\end{sc}
\end{small}
\end{center}
\vskip -0.1in
\end{table*}

\begin{table*}[t]
\caption{Mean and standard deviation of accuracy, AUC, sensitivity and specificity for different train-test split using standard DCNS on Pulmonary Edema. It cannot be said that the deeper the model the better the performance is as ResNet-50 shoiws consistently better performance.}

\label{PulAtl}
\vskip 0.15in
\begin{center}
\begin{small}
\begin{sc}
\begin{tabular}{lcccccr}
\hline
 & Accuracy (\%) & AUC & Sensitivity (\%) & Specificity (\%) \\
\hline
Alex Net    & $85.33 \pm 5.81\%$ & $0.87 \pm 0.06$  & $74.66 \pm 12.90\%$ & $88.00 \pm 07.65\%$ \\
VGG-16      & $89.33 \pm 5.94\%$ & $0.92 \pm 0.05$  & $87.11 \pm 10.22\%$ & $91.55 \pm 08.90\%$ \\
VGG-19      & $88.89 \pm 4.48\%$ & $0.93 \pm 0.06$  & $87.56 \pm 06.10\%$ & $90.22 \pm 07.07\%$ \\
ResNet-50   & $\bf90.22 \pm 3.67\%$ & $\bf0.94 \pm 0.04$  & $\bf88.89 \pm 06.51\%$ & $91.56 \pm 06.89\%$ \\
ResNet-101  & $86.88 \pm 6.95\%$ & $0.90 \pm 0.07$  & $85.33 \pm 08.80\%$ & $88.44 \pm 11.12\%$ \\
ResNet-152  & $87.55 \pm 5.56\%$ & $0.91 \pm 0.06$  & $82.67 \pm 10.02\%$ & $\bf92.44 \pm 08.68\%$ \\

\hline
\end{tabular}
\end{sc}
\end{small}
\end{center}
\vskip -0.1in
\end{table*}

\begin{figure}[t]
\centering
\includegraphics[width=7.5cm, height=10cm]{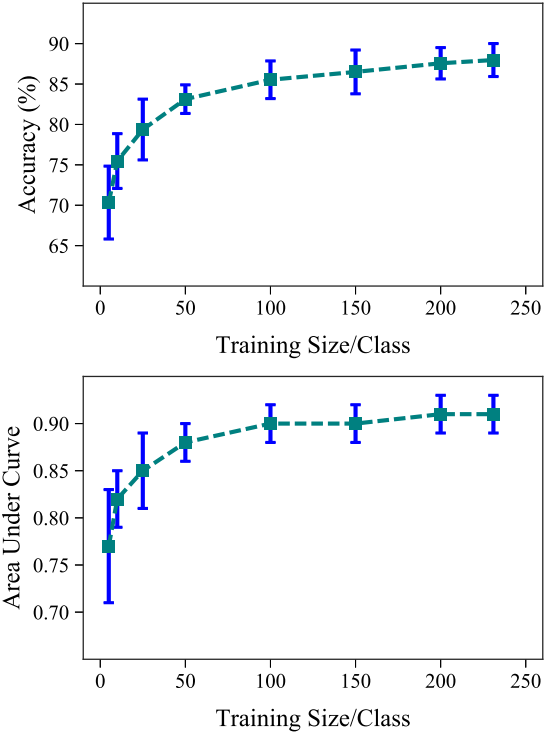}\\
\caption{Accuracy (top) and AUC (bottom) of Cardiomegaly detection increases as number of samples in training set is increased. By $200$ samples both accuracy and AUC saturates.} 
\label{fig:traing_size_card}
\end{figure}

Having around 600 images for training a network is not sufficient. We wanted to see how does the mean accuracy and the standard deviation vary as we change the number of training examples. Since averaging over multiple train-test splits gave a robust classification accuracy and other figures of merit, we used this classification process to identify the deviation of the result as a function of the number of training images. The results are shown in Fig.~\ref{fig:traing_size_card}. As expected, for both accuracy and AUC, the mean is lower and deviation is higher for less than 50 training example per category. As the number of example increases, the mean increases and the deviation decreases coming to a saturation at about 200 images.

To check if the same model gives the highest accuracy for different abnormalities, we model pulmonary edema using the same averaging process described above. Our dataset for the detection of pulmonary edema contains available 45 frontal CXR images with pulmonary edema and randomly chosen 45 normal frontal CXRs from the Indiana Dataset. We partitioned the dataset in train and test set such that 30 of each class have been selected for training and 15 of each class for testing. We have run our program with 15 different seeds and reported the overall performance metrics in the table \ref{PulAtl} as (mean {$\pm$} s.d.). We find that whereas ResNet-152 gave the highest accuracy for cardiomegaly detection, for pulmonary edema detection ResNet-50 gives the highest accuracy, highest AUC and highest sensitivity. ResNet-152 has a slightly higher specificity. This shows that there is no single model appropriate for all abnormalities, rather the suitable network varies for different abnormalities. This observation is consistent with the conclusions drawn in \cite{azizpour2015generic}. In this case, ResNet-152 which gave the highest accuracy for cardiomegaly detection achieves almost one percentage point reduced accuracy compared to ResNet-50.

\subsubsection{Classification using ensemble of models}

\begin{figure*}[t]
\centering
\includegraphics[width=16cm, height=8cm]{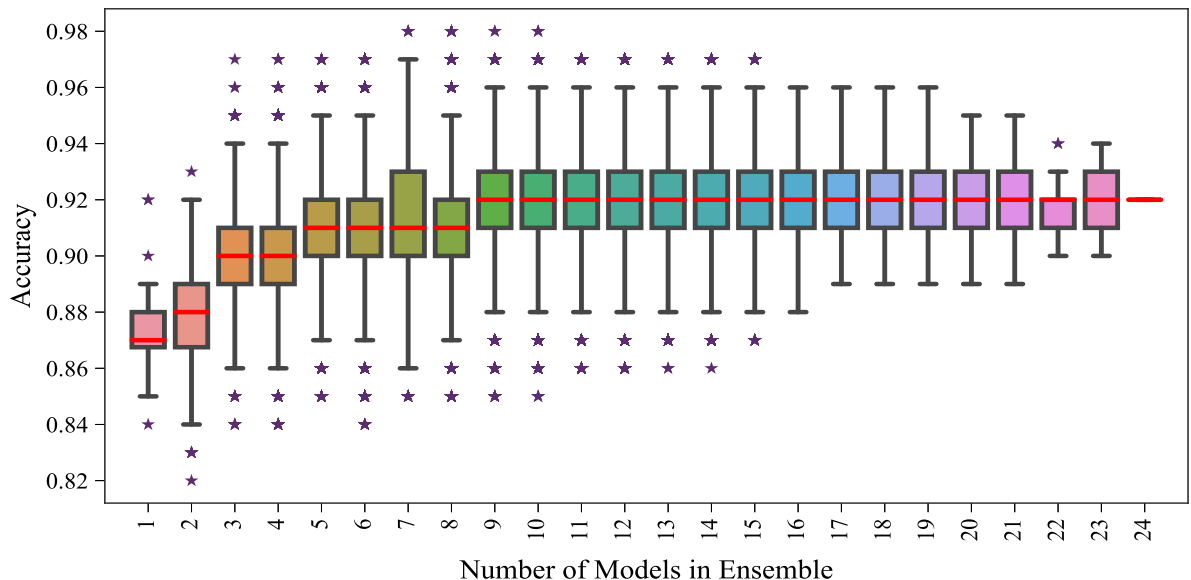}\\
\caption{Ensemble of all the combinations of the trained models using linear averaging. As the number of models in the ensemble increases the performance become robust.} \label{fig:ensemble_card}
\end{figure*}

We trained four different instances of each of the DCNs, i.e, AlexNet, VGG-16, VGG-19, ResNet-50, ResNet-101 and ResNet-152, to detect cardiomegaly. Thus a total of $24$  networks were trained on the same training data. There are a number of ways to perform ensemble on the trained model. The methods include linear averaging, bagging, boosting, stacked regression \cite{breiman1996stacked} etc. Since, the number of images in the training dataset is only 564, which is far less than the number of trainable parameters in the classifiers, the individual classifiers always overfit the training set. In this situation, if bagging, boosting and/or stacked regression are employed to build the ensemble model, it will result in a completely biased model. Thus, the ensemble models were obtained by using simple linear averaging of the probabilities given by the individual models. The performance of the ensembles was measured using 50 cardiomegaly and 50 normal images for all the possible combinations of the trained individual models. The performance of these combinations is shown in Fig. \ref{fig:ensemble_card} using boxplots. The horizontal red bars indicate the 50 percentile values and the spread of the blue boxes indicate the 25 and 75 percentile values. The black stars indicate extreme points in the data. It can be observed from the figure that, combinations of 7 to 10 models can achieve higher accuracy, however they have the largest spread. On the other hand, as number of models in the ensemble increases, the accuracy of the ensemble model converges to a certain value which for this experiment was $92\%$. 

\begin{figure}[tbh]
\centering
\includegraphics[width=8cm, height=6cm]{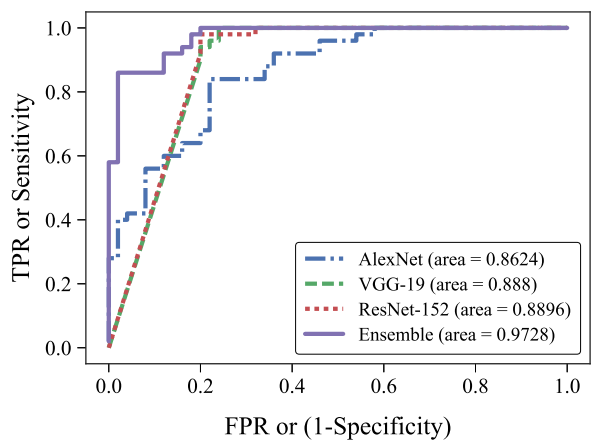}\\
\caption{ROC Curve of AlexNet, VGG-19, ResNet-152 and an ensemble model. The ensemble model shows increased area under the curve compared to individual models.} \label{roc_max_card}
\end{figure}

The ROC curves of one instance AlexNet, VGG-19, ResNet-152 and one ensemble model, that is linear average of 6 different types of DCNs, are shown in Fig. \ref{roc_max_card}. The curves are obtained using 50 cardiomegaly and 50 normal images. The AUC obtained for each model are $0.8624$, $0.8888$, $0.8896$ and $0.9728$, respectively. We can understand from the AUC values that, the separation between the pathology class and the normal class increases when an ensemble of multiple DCNs are performed. For the ensemble model to be used as a screening tool with high sensitivity, the operating point on the curve is set to achieve $98\%$ sensitivity. The specificity obtained at this point is $82\%$. The second operating point is set for high specificity of $98\%$ and the sensitivity at this point is $86\%$.


\subsection{Localization}
\label{results_localization}
For any diagnostic task, it is desirable to gain intuitive understanding of why a certain classification decision is made rather than being a black box method. In other words, it is desirable to distinguish features that contributed most to certain abnormality in the entire chest X-Ray. There are various ways of achieving this goal \cite{zeiler2014visualizing, bach2015pixel, springenberg2014striving, zhou2016learning}. The method used in \cite{zeiler2014visualizing} is the simplest, where a patch is occluded in the image to measure its impact on the eventual classification confidence score. We have used this method to find the regions in the image responsible for a certain abnormality detection. As a representative example, we have used cardiomegaly and pulmonary edema which occur in heart and lung areas respectively. The localization scheme described in section \ref{sec:loc_scheme} is followed with a patch size of $40\times40$ pixels taking lowest $20\%$ values of probabilities. Instead of gray level occlusion as in \cite{zeiler2014visualizing} we found that black level occlusion works better for CXRs. This is due to the fact that the CXRs themselves are mostly gray level and occlusion of the same level does not hide much information compared to the neighborhood.

\subsubsection{Cardiomegaly Localization}
\begin{figure}[!tbh]
\begin{minipage}[b]{0.49\linewidth}
  \centering
  \centerline{\includegraphics[width=3.45cm]{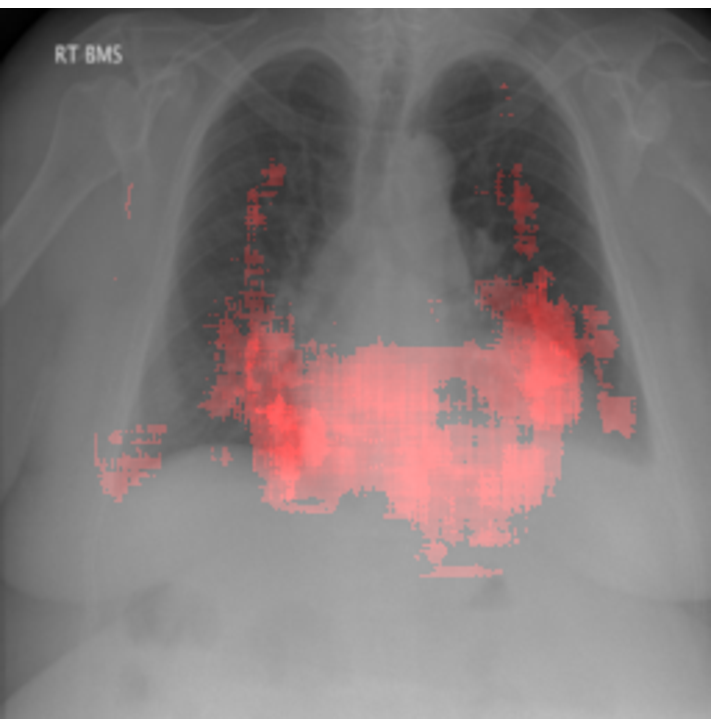}}
  \centerline{\footnotesize{(a)}}\medskip
\end{minipage}
\hfill
\begin{minipage}[b]{0.49\linewidth}
  \centering
  \centerline{\includegraphics[width=3.45cm]{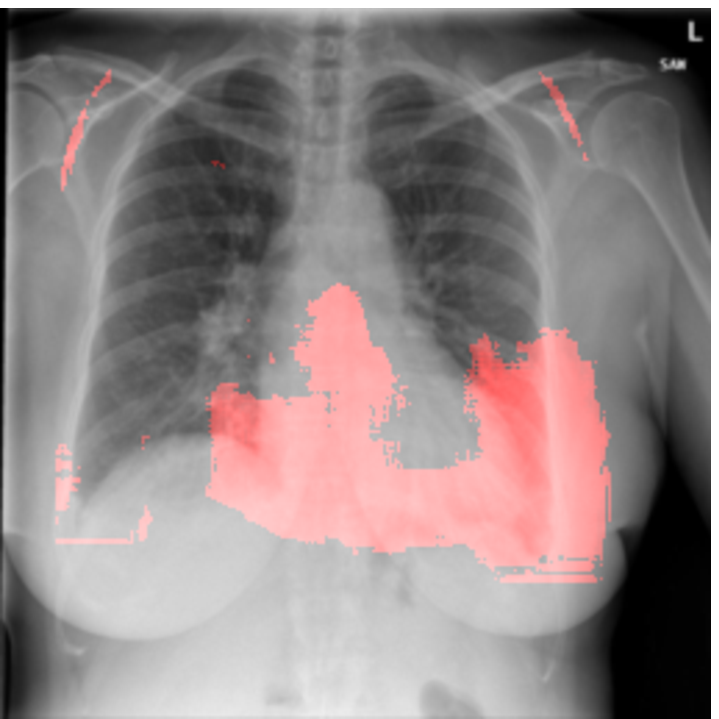}}
  \centerline{\footnotesize{(b)}}\medskip
\end{minipage}
\caption{The localization observed for Cardiomegaly. The localized area by the probability map is superimposed on the CXRs and localization around the heart is observed. This observation is counter intuitive to the rule based method which involves relative size of heart and lung.} \label{fig:local_card}
\end{figure}

The localization of abnormalities in cardiomegaly examples are shown in Fig. \ref{fig:local_card}. Here, $20\%$ of the image area is shown which has the highest sensitivity. It can be observed from the figures that the network is indeed most sensitive to the region where the heart is larger than a normal heart. We have performed this experiment on $50$ cardiomegaly and $50$ normal images and found this localization to be consistent for most examples. There is not much functional difference between a normal and cardiomegaly example other than the fact that the heart in cardiomegaly is larger than a normal heart. Given the fact that the normal images could also have various size of heart depending on the age or physical attributes of a patient, we found this level of localization sensitivity to be remarkable. Also interesting is the fact that the standard rule based features like CTR and CTAR take into account the relative size of heart and lung to determine if there is cardiomegaly present or not. In the DCN localization experiment, we see counter-intuitively that most of the signals contributing to the softmax score are coming from the heart only. This means that there are characteristic features in the shape of the heart and its surrounding regions that alone is sufficient to detect cardiomegaly. The lung and its relative size are probably less important features when trying to detect cardiomegaly. This observation is counterintuitive and needs to be explored further in future work.  

\subsubsection{Pulmonary Edema Localization}
\begin{figure}[t]
\begin{minipage}[b]{0.49\linewidth}
  \centering
  \centerline{\includegraphics[width=3.45cm]{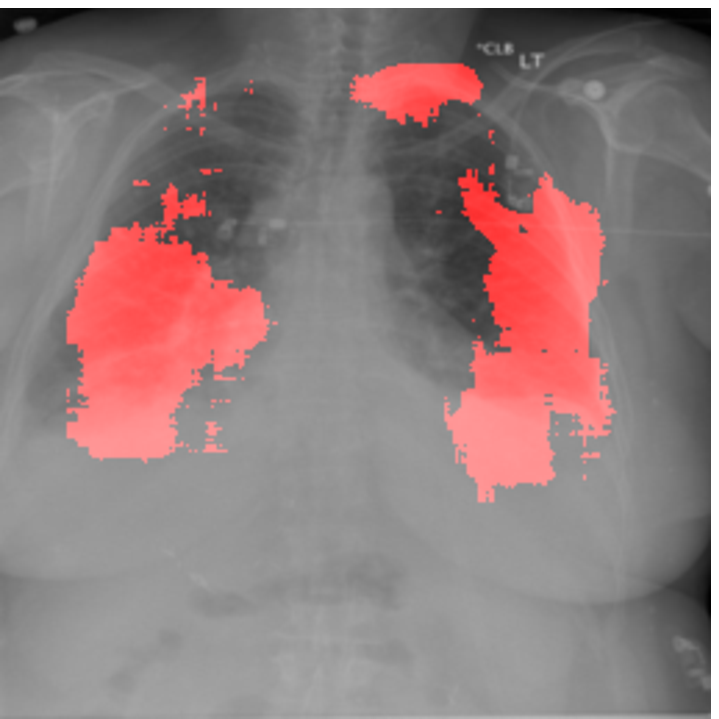}}
  \centerline{\footnotesize{(a)}}\medskip
\end{minipage}
\hfill
\begin{minipage}[b]{0.49\linewidth}
  \centering
  \centerline{\includegraphics[width=3.45cm]{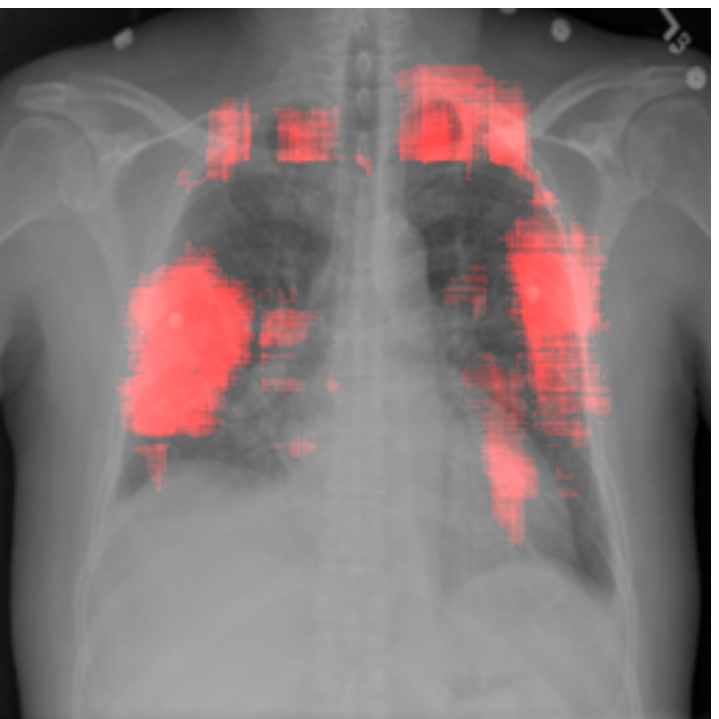}}
  \centerline{\footnotesize{(b)}}\medskip
\end{minipage}
\caption{The localization observed for pulmonary edema. The localized area by the probability map is superimposed on the CXRs and localization around excess fluid in lungs is observed.}\label{fig:local_pedema}
\end{figure}

In order to test the effectiveness of the localization procedure in  areas other than the heart region, we chose pulmonary edema which occurs in the lung region. Also, pulmonary edema is detected by the net like white structure in the lung area. No anatomical shape change is associated with the abnormality. We have found that the localization is obtained best when the ROIs of lungs are taken to compute the map. Following the scheme in section \ref{sec:loc_scheme}, localization experiment on pulmonary edema is performed as shown in Fig. \ref{fig:local_pedema}. It has been observed that the classifier is not sensitive to the fine features like septal or Kerley B lines. The localization is mainly obtained in the lung region where excess fluid is observed. Some localization regions are outside the lung region which occurs primarily for the fact that, even though the occlusion center is outside the lung, it occludes lung region and thus the probability drop occurs. 


\subsection{Comparison between Rule based and DCN based cardiomegaly detection} \label{card_describe}

\begin{table*}[htbp]
\caption{Accuracy, AUC, sensitivity and specificity for rule based methods, standard DCNs and ensemble of DCNs for Cardiomegaly detection. The ensemble model provides highest accuracy and AUC.} 
\label{RulevDeep}
\vskip 0.15in
\begin{center}
\begin{small}
\begin{sc}
\begin{tabular}{lcccccr}
\hline
 & Accuracy (\%) & AUC & Sensitivity (\%) & Specificity (\%) \\
\hline
Candemir~\cite{candemir2016automatic} [\ldots]   & $76.50\%$ & $-$ & $77.10\%$ & $76.40\%$\\
Rule based features    & $75.60\%$ & $0.80$ & $73.60\%$ & $77.60\%$\\
Alex Net    & $86.00\%$ & $0.92$  & $86.00\%$ & $86.00\%$ \\
VGG-16      & $86.00\%$ & $0.87$  & $\mathbf{96.00\%}$ & $76.00\%$ \\
VGG-19      & $92.00\%$ & $0.94$  & $92.00\%$ & $92.00\%$ \\
ResNet-50   & $87.00\%$ & $0.93$  & $94.00\%$ & $80.00\%$\\
ResNet-101  & $92.00\%$ & $0.92$  & $88.00\%$ & $\bf96.00\%$\\
ResNet-152  & $90.00\%$ & $0.91$  & $92.00\%$ & $88.00\%$\\
Ensemble	& $\mathbf{93.00\%}$ & $\mathbf{0.97}$ & $94.00\%$ & $92.00\%$\\

\hline
\end{tabular}
\end{sc}
\end{small}
\end{center}
\vskip -0.1in
\end{table*}

A comparison between rule based and DCN based cardiomegaly detection is shown in table~\ref{RulevDeep}. State-of-the-art method by Candemir \textit{et al.}~\cite{candemir2016automatic} reported an accuracy of $76.5\%$ while classifying between 250 cardiomegaly and 250 normal images. They employed 1D-CTR, 2D-CTR, and CTAR computed from segmented CXRs as features. A brief discussion about the rule based approach is given in the supplementary materials in section \ref{sec:rule_based}. In verifying that claim in the paper, we reproduced those results and achieved an accuracy of $75.6\%$ on the same train-test set split on which the DCNs are trained. It can be observed from the table that the results are similar to that obtained by~\cite{candemir2016automatic}. However, it is evident from the table that all DCN based approaches outperform the rule based method. As stated earlier, DCNs were fine tuned on a sample of 560 images and validated on 100 images. Among the independent DCN models, VGG-19 model achieves the highest accuracy of $92\%$ and highest AUC of $0.9408$ for detecting cardiomegaly. The ensemble model, which is linear average of the six individual DCN models, shows the best accuracy of $93\%$ and AUC of $0.9728$. The accuracy is $17$ percentage point higher than  that reported in Candemir's paper and the AUC is $18$  percentage points higher than our implementation of the paper. Similarly, a $17$ percentage higher sensitivity and $16$ percentage point higher specificity from the Candemir's paper is reported. This quantum of improvement in accuracy, AUC, sensitivity and specificity makes a strong case for use of deep learning based detection techniques in real world application of medical image analysis.


\subsection{Tuberculosis Detection} \label{tb_describe}
In this section we evaluate the effectiveness of the network design and DCN pipelines for a different dataset and abnormality. We use the Shenzen dataset as it is often used for reporting accuracy on tuberculosis detection. 
\begin{table*}[htbp]
\caption{Accuracy, AUC, sensitivity and specificity for different methods for Tuberculosis detection using Shenzhen Dataset. The ensemble model provides highest accuracy and AUC.} 
\label{RulevDeepTB}
\vskip 0.15in
\begin{center}
\begin{small}
\begin{sc}
\begin{tabular}{lcccccr}
\hline
 & Accuracy (\%) & AUC & Sensitivity (\%) & Specificity (\%) \\
\hline
Jaeger~\cite{jaeger2014automatic} [\ldots] & $84.10\%$ & $0.90$ & $-$ & $-$\\
Hwang~\cite{hwang2016novel} [\ldots]  & $83.70\%$ & $0.93$ & $-$ & $-$  \\
Lopes and Valiati~\cite{lopes2017pre} [\ldots] & $84.60\%$ & $0.93$ & $-$ & $-$ \\
Alex Net    & $84.00\%$ & $0.89$ & $72.00\%$ & $\bf 96.00\%$ \\
VGG-16      & $84.00\%$ & $0.88$ & $\bf 96.00\%$ & $72.00\%$ \\
VGG-19      & $80.00\%$ & $0.89$ & $76.00\%$ & $84.00\%$ \\
ResNet-50   & $86.00\%$ & $0.90$ & $84.00\%$ & $88.00\%$\\
ResNet-101  & $84.00\%$ & $0.85$ & $88.00\%$ & $80.00\%$\\
ResNet-152  & $88.00\%$ & $0.91$ & $80.00\%$ & $92.00\%$\\
Ensemble	& $\bf 90.00\%$ & $\bf 0.94$ & $88.00\%$ & $92.00\%$\\

\hline
\end{tabular}
\end{sc}
\end{small}
\end{center}
\vskip -0.1in
\end{table*}
Detailed study of tuberculosis detection will be provided in a future publication. But the a comparison among several TB classification methods and proposed DCN based methods along with their ensemble using Shenzhen Dataset is shown in table~\ref{RulevDeepTB}. Previously, Jaeger~\emph{et.~al}~\cite{jaeger2014automatic} extracted several features from lung segmented CXRs and employed various classification methods to benchmark the features. The results reported in the table is obtained using low-level content-based image retrieval based features and linear logistic regression based classification.
Hwang~\emph{et.~al}~\cite{hwang2016novel} trained three different DCNs on three different train/test split on a large private KIT dataset and tested the ensemble of the model on Shenzhen dataset. It is to be noted that, both the KIT and Shenzhen dataset were obtained using Digital Radiography.
Lopes and Valiati~\cite{lopes2017pre} employed bags of features and ensemble method using features from ResNet, VGG and GoogLeNet models and trained SVM classifier on them. They obtained highest AUC in Shenzhen dataset using ensemble of individual SVM classifiers.
Lakhani and Sundaram~\cite{lakhani2017deep} employed AlexNet and GoogLeNet on a combined dataset of four different databases and performed ensemble on the trained models. They do not report test results on Shenzhen dataset and thus it was not shown in the table. 
The DCN based methods shown in table~\ref{RulevDeepTB} have comparable or higher accuracy and lower AUC than the results already present in the literature. The VGG-16 model obtains highest sensitivity and AlexNet model obtains highest specificity. The sensitivity and specificity measures for Jaeger's, Hwang's and Lopes and Valiati's paper are not shown in the table as they were not reported in the respective papers.
In terms of accuracy and AUC, our ensemble method obtains highest values of $90\%$ and $0.94$, respectively. This accuracy is obtained when classifier threshold is set to $0.74$. When classifier threshold is set to $0.50$, the accuracy obtained is $88\%$. Thus, we report a $5$ percentage point higher accuracy and $1$ percentage point higher AUC compared to nearest Lopes and Valiati's paper.


\section{Conclusion}
\label{conclusion}
 
In summary, we have explored DCNN based abnormality detection in frontal chest XRays. We have found the existing literature to be insufficient for making comparison of various detection  techniques either due to studies reported on private datasets or not reporting the test scores in proper detail~\cite{shin2016learning}. In order to overcome these difficulties, we have used the publicly available Indiana chest X-Ray dataset and studied the performance of various DCN architectures on different abnormalities. We have found that the same DCNN architecture doesn't perform well across all abnormalities. When the number of training examples is low, a consistent detection result can be achieved by doing multiple train-test with random data split and the average values are used as the accuracy measure. Shallow features or earlier layers  consistently provide higher detection accuracy compared to deep features. We have also found ensemble models to improve classification significantly compared to single model when only DCNN models are used. Combining DCNN models with rule based models degraded the accuracy. Combining these insights, we have reported the highest accuracy on a number of chest X-Ray abnormality detection where comparison could be made. For the cardiomegaly classification task, the deep learning method improves the accuracy by a staggering 17 percentage point. Using the same method developed in the paper, we achieve the highest accuracy on the Shenzen dataset for Tuberculosis detection. We have also performed localization of features responsible for classification decision. We found that for spatially spread out abnormalities like cardiomegaly and pulmonary edema, the network can localize the abnormalities successfully most of the time. However, the localization fails for pointed features like lung nodule or bone fracture. One remarkable result of the cardiomegaly localization is that the heart and its surrounding region is most responsible for cardiomegaly detection. This is counterintuitive considering the usual method of using the ratio of heart and lung area as a measure for cardiomegaly. However, expert radiologists often conclude upon cardiomegaly by looking at the heart's shape rather than using a quantitative method. We believe that through deep learning based classification and localization, we will discover many more interesting features that are not considered traditionally. 

While finishing this paper, we became aware of a new dataset announcement and paper focused on similar problem~\cite{wang2017ChestX-ray8}. It would be interesting to apply the techniques discussed our paper on the new dataset becomes available.

\section{Acknowledgement}
\label{acknowledge}
This research used resources of the National Energy Research Scientific Computing Center, a DOE Office of Science User Facility supported by the Office of Science of the U.S. Department of Energy under Contract No. DE-AC02-05CH11231. Thanks Leonid Oliker at NERSC for sharing his allocation on the OLCF Titan supercomputer with us on project CSC103. Thanks to Hoo-Chang Shin for correspondence regarding the Indiana chest X-Ray dataset.

\bibliography{refs}
\bibliographystyle{IEEEbib}

\clearpage

\section{Supplementary Materials}~\label{supplimentaryMaterials}
In this section, first a detailed description about the segmentation of the CXRs is given. Then the rule based machine learning model is described. After that we show classification results of 20 chest X-Ray abnormalities. And finally we show more localization results with additional insights.

\subsection{Rule based Approach in Detecting Cardiomegaly}\label{sec:rule_based}

This method uses existing CXRs and their radiologist marked lung/heart boundaries as models, and estimates the lung/heart boundary of a patient X-ray by registering the model X-rays to the patient X-ray.~\cite{candemir2016automatic}

\subsubsection{Using Radon Transform and Bhattacharyya Distance to find visually similar images}

\begin{figure}[h]
\centering
\includegraphics[width=7.5cm, height=6cm]{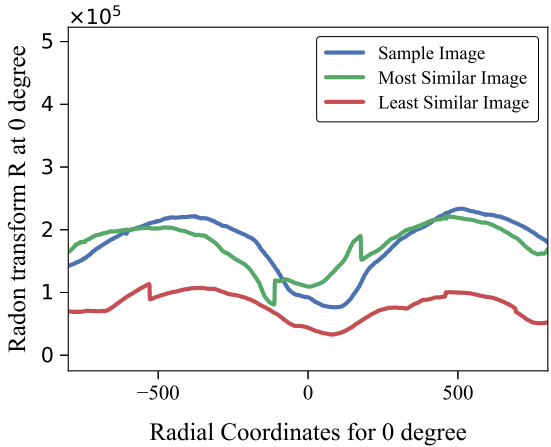}\\
\caption{Comparison of Radon Transforms among sample CXR, the most similar and the least similar CXR at 0 Degree}
\end{figure}

We use the publicly available chest x-ray dataset (JSRT)~\cite{shiraishi2000development} with reference boundaries given in the SCR dataset~\cite{van2006segmentation}. For a given test image, the radon transform of that image is calculated at radial coordinates, ranging from 0 to 90 degree. The radon function computes projections of an image along specified directions. Bhattacharyya distance~\cite{bhattacharyya1943measure} is calculated between radon transform of test CXR and the sample CXRs to find 5 visually similar samples from the JSRT dataset. We use the most similar images from the dataset to register to the test image. As mentioned by Candemir \textit{et al.}~\cite{candemir2016automatic}, the main objective of similarity measurement is to increase the correspondence performance and reduce the computational cost during registration.

\subsubsection{Calculating correspondence between test CXR and model CXRs using SIFTFlow}

We compute the correspondence map between the test CXR and visually similar CXR models by calculating local image features and matching the most similar locations. We employ the SIFT-flow algorithm which matches densely sampled SIFT features between two images. The computed correspondence map is a transformation from model X-ray to the patient X-ray. Finally, the computed transformation matrix is applied on the model CXR's lung-heart boundary to generate an approximate lung-heart segmentation of the test image.

\begin{figure}[t!]
\centering
 \begin{minipage}[t!]{0.22\textwidth}
    \includegraphics[width=3.8cm, height=3.8cm]{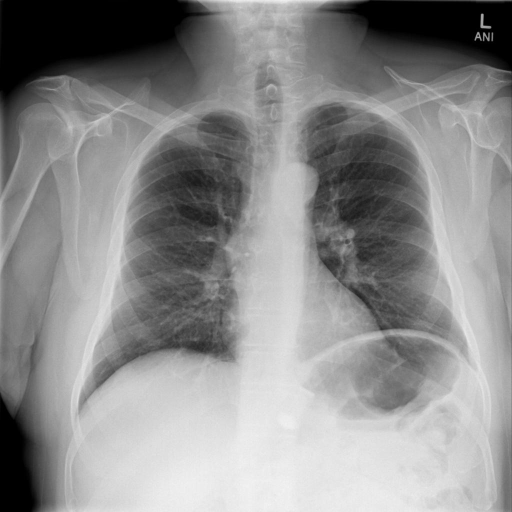}
    \centerline{\footnotesize{(a)}}\medskip
  \end{minipage}
  \hfill
 \begin{minipage}[t!]{0.22\textwidth}
    \includegraphics[width=3.8cm, height=3.8cm]{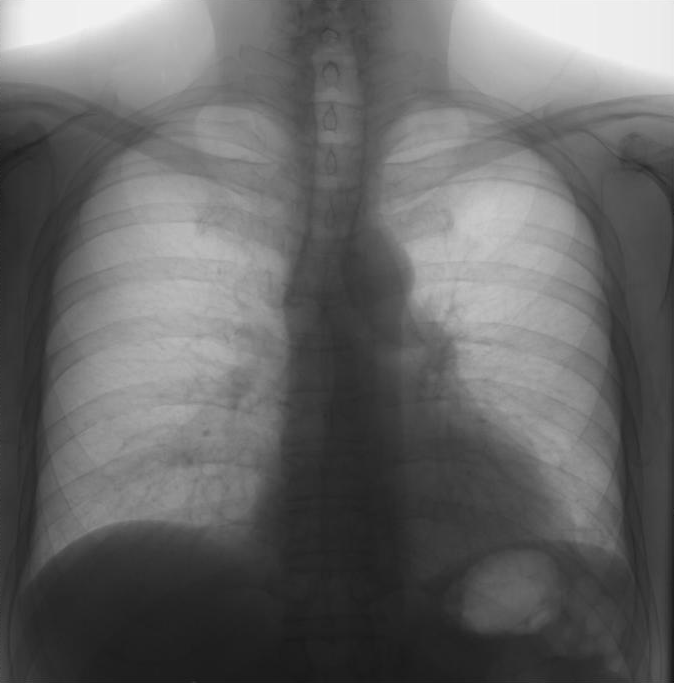}
    \centerline{\footnotesize{(b)}}\medskip
  \end{minipage}
  \hfill
  \begin{minipage}[t!]{0.22\textwidth}
    \includegraphics[width=3.8cm, height=3.8cm]{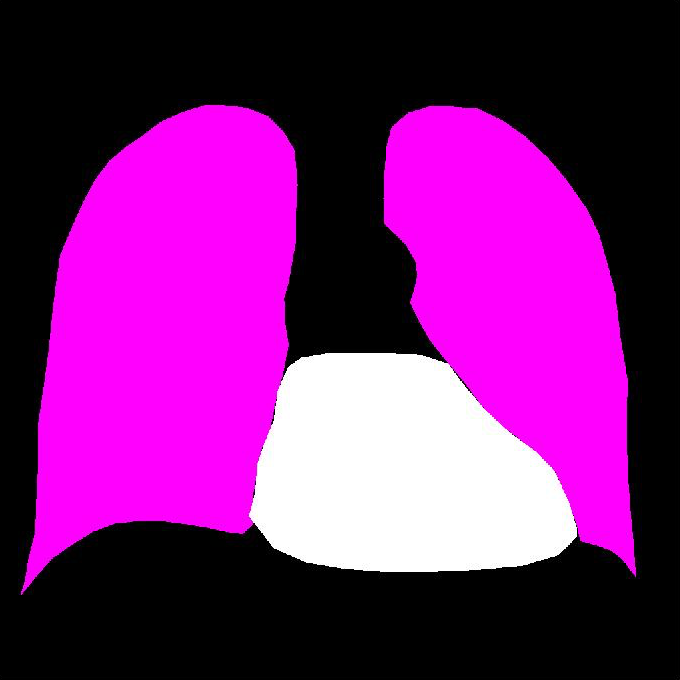}
    \centerline{\footnotesize{(c)}}\medskip
  \end{minipage}
  \hfill
  \begin{minipage}[t!]{0.22\textwidth}
    \includegraphics[width=3.8cm, height=3.8cm]{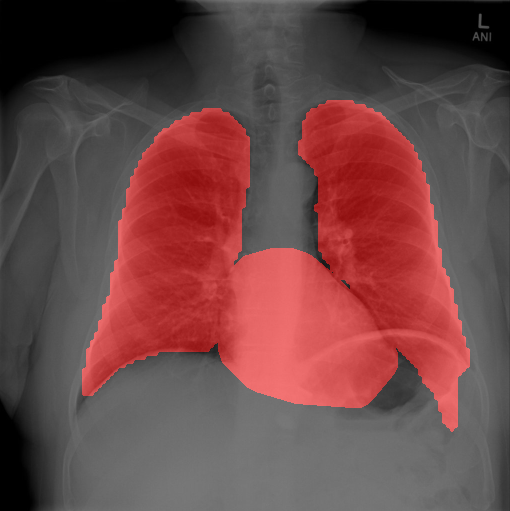}
    \centerline{\footnotesize{(d)}}\medskip
    \end{minipage}
  \caption{(a) Test CXR, (b) Sample Similar CXR from the JSRT dataset, (c) Sample CXR's lung and heart model atlas, (d) Heart and lung segmentation of test CXR.}
\end{figure}

\subsubsection{Rule based feature extraction}
Rule based features are extracted using \ref{1D-CTR}, \ref{2D-CTR} and \ref{CTAR} and SVM was used to classify between cardiomegaly and normal CXR images.

\subsection{Classification Results on the 20 chest X-Ray abnormalities}

\begin{table*}[htbp]
\caption{Performance analysis of detecting 20 abnormalities in CXR with ResNet-152.Experiments were performed by taking same number of normal images and then split images into train test sets with 70:30 ratio for all abnormalities.}
\label{20abnormality}
\vskip 0.15in
\begin{center}
\begin{small}
\begin{sc}
\begin{tabular}{lcccccr}
\hline
Abnormalities & Total images & Accuracy (\%) & AUC & Sensitivity (\%) & Specificity (\%) \\
\hline
Calcified Granuloma  & 314 & 64.74\% & 0.65 & 61.05\% & 68.42\% \\
Pulmonary Atlectasis  & 313 & 80.00\% & 0.87 & 68.75\% & 91.25\% \\
Calcinosis & 288 & 68.39\% & 0.79 & 45.98\% & 90.80\% \\
Granulomatous Disease  & 162 & 68.37\% & 0.71 & 61.22\% & 75.51\% \\
Pleural Effusion  & 147 & 85.56\% & 0.89 & 86.67\% & 84.44\% \\
Atherosclerosis  & 127 & 79.49\% & 0.84 & 82.05\% & 76.92\% \\
Airspace Disease  & 112 & 79.41\% & 0.78 & 94.12\% & 64.71\% \\
Nodule  & 106 & 67.19\% & 0.69 & 46.88\% & 87.50\% \\
Pulmonary Emphysema  & 103 & 91.94\% & 0.96 & 93.55\% & 90.32\% \\
Scoliosis & 94 & 77.59\% & 0.83 & 68.97\% & 86.21\% \\
Fractures Bone  & 88 & 79.63\% & 0.78 & 88.89\% & 70.37\% \\
Osteophyte  & 69 & 73.81\% & 0.76 & 85.71\% & 61.90\% \\
Pulmonary Congestion  & 68 & 95.24\% & 0.97 & 100.00\% & 90.48\% \\
Bullous Emphysema  & 66 & 82.50\% & 0.87 & 75.00\% & 90.00\% \\
Subcutaneous Emphysema  & 64 & 87.50\% & 0.91 & 85.00\% & 90.00\% \\
Spondylosis  & 64 & 67.50\% & 0.65 & 80.00\% & 55.00\% \\
Emphysema  & 62 & 86.84\% & 0.94 & 84.21\% & 89.47\% \\
Granuloma  & 52 & 68.75\% & 0.66 & 93.75\% & 43.75\% \\
Hernia Hiatal  & 46 & 89.29\% & 0.90 & 92.86\% & 85.71\% \\
Pulmonary Disease Chronic Obstructive  & 46 & 85.71\% & 0.87 & 92.86\% & 78.57\% \\
\end{tabular}
\end{sc}
\end{small}
\end{center}
\end{table*}

In this section, we report the classification accuracy, sensitivity and specificity using the ResNet-152 model. We hope that these numbers will set a benchmark to compare against other machine learning methods on this dataset.

\section{Additional Examples of Localization}

\begin{figure}[h]
\centering
\includegraphics[width=3.5cm, height=3.5cm]{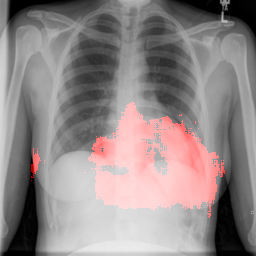}
\includegraphics[width=3.5cm, height=3.5cm]{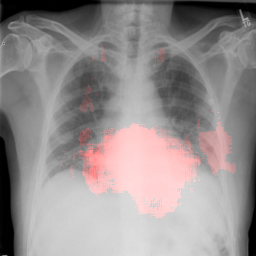}\\
(a)\hspace{3.5cm}(b)\\
\includegraphics[width=3.5cm, height=3.5cm]{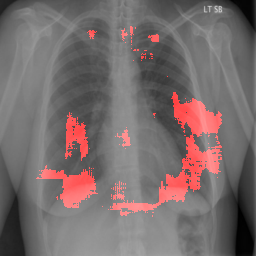}
\includegraphics[width=3.5cm, height=3.5cm]{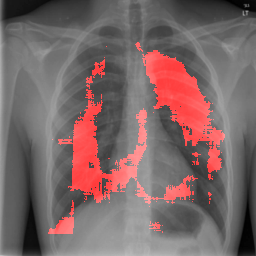}\\
(c)\hspace{3.5cm}(d)\\
\includegraphics[width=3.5cm, height=3cm]{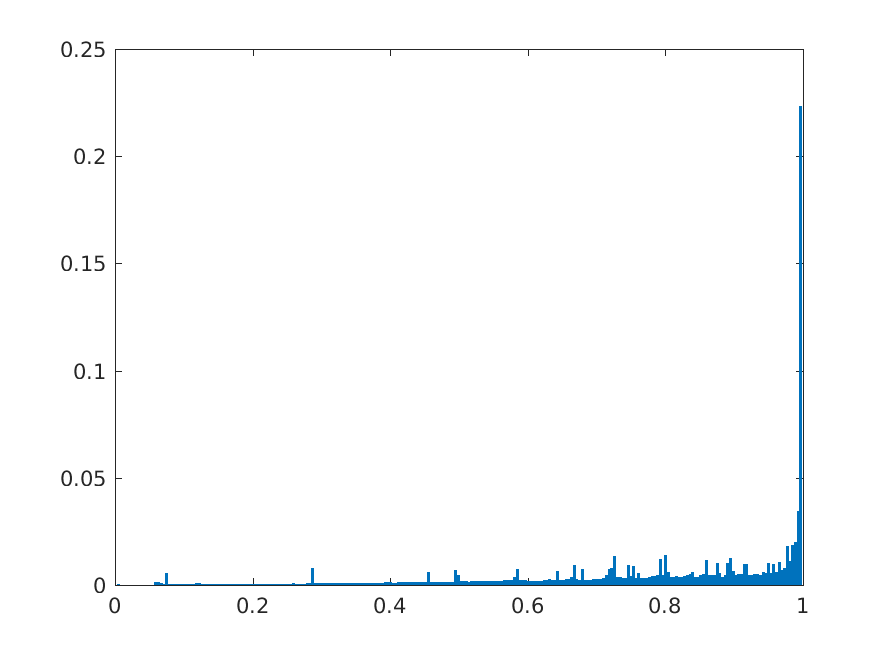}
\includegraphics[width=3.5cm, height=3cm]{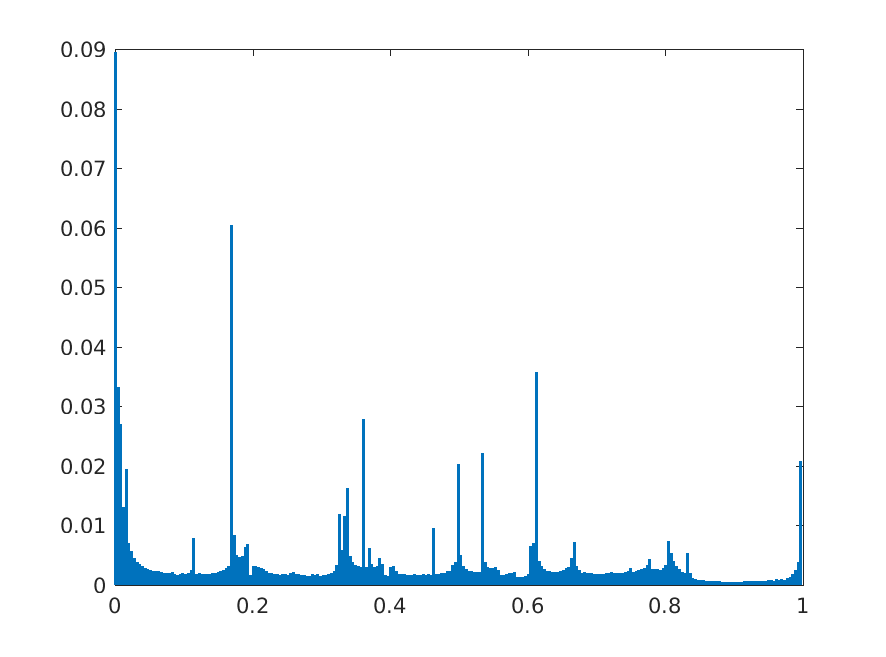}\\
(e)\hspace{3.5cm}(f)\\
\caption{(a) (b) Fine localization around the heart for Cardiomegaly detection in CXRs with cardiomegaly. (c) (d) Spread out localization for Cardiomegaly detection in normal CXRs. (e) Average histogram of heat maps of CXRs with Cardiomegaly showing high sensitivity toward the disease. (f) Average histogram of heat maps for Cardiomegaly detection in normal CXRs showing a relatively spread out distribution.}\label{fig:more_card_loc}
\end{figure}

\begin{figure}[h]
\centering
\includegraphics[width=3.5cm, height=3.5cm]{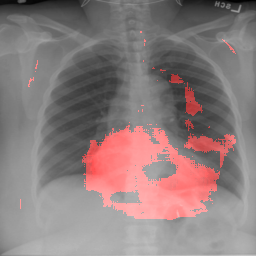}
\includegraphics[width=3.5cm, height=3.5cm]{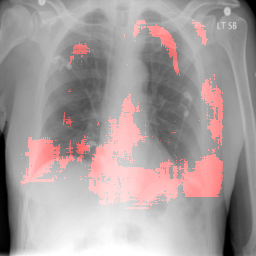}\\
(a)\hspace{3.5cm}(b)\\
\caption{Some examples of failure cases of Cardiomegaly classification. (a) Localization on a CXR with Cardiomegaly which the classifier detects as normal. (b) Localization on a normal CXR where the classifier detects Cardiomegaly.}\label{fig:card_loc_failures}
\end{figure}

In this section we show more examples of localization. Few localization samples are shown in Fig. \ref{fig:more_card_loc}. It can be observed that, in the CXRs with Cardiomegaly (Fig. \ref{fig:more_card_loc}(a) and (b)) a fine localization around the heart is observed. In the normal CXRs (Fig. \ref{fig:more_card_loc}(c) and (d)) such localization is not observed. Rather the lowest $20\%$ probabilities are spread out in the CXR image. It is interesting to note that, the localization algorithm gets low probability where the heart is enlarged during cardiomegaly, but the proportion is small compared to the localization in other areas of normal CXRs. In order to observe the performance of the heat map we computed histograms of heat maps of each of the 100 CXRs in the test set for Cardiomegaly detection and average histograms are shown in Fig. \ref{fig:more_card_loc}(e) and (f) for CXRs with Cardiomegaly and normal CXRs, respectively. It is to be noted that, the histograms include both success and failure cases. It can be observed that, for CXRs with Cardiomegaly the classifier is highly sensitive toward Cardiomegaly detection even under occlusion. This indicates that, the classifier primarily looks for local features in a CXR instead of some feature that is spread out in the entire CXR. However, the classifier is not sensitive toward normal CXRs under occlusion. Rather, the probabilities are spread out in the probability spectrum. After that, we analyzed the failure cases where the classifier is unable to classify the image correctly. Two such examples of failure cases are shown in Fig. \ref{fig:card_loc_failures}. The localized CXR shown in Fig. ~\ref{fig:card_loc_failures}(a) contains Cardiomegaly whereas the classifier detects it as normal. However, the localization shows that it localizes around heart quite well despite the in accurate classification. On the other hand, Fig. \ref{fig:card_loc_failures}(b) shows an example of normal image which has been classified as Cardiomegaly by the classifier. There is stronger localization around the hear that that is observed for normal images as in Fig. \ref{fig:more_card_loc}(c) and (d), however, like those images the localization is spread out.

\begin{figure}[h]
\centering
\includegraphics[width=3.5cm, height=3.5cm]{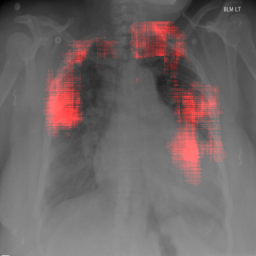}
\includegraphics[width=3.5cm, height=3.5cm]{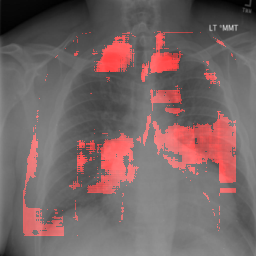}\\
(a)\hspace{3.5cm}(b)\\
\includegraphics[width=3.5cm, height=3.5cm]{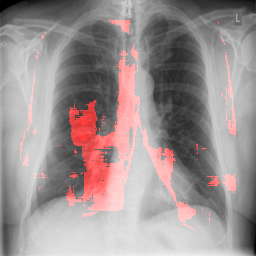}
\includegraphics[width=3.5cm, height=3.5cm]{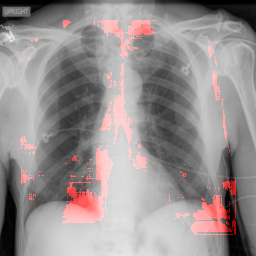}\\
(c)\hspace{3.5cm}(d)\\
\includegraphics[width=3.5cm, height=3cm]{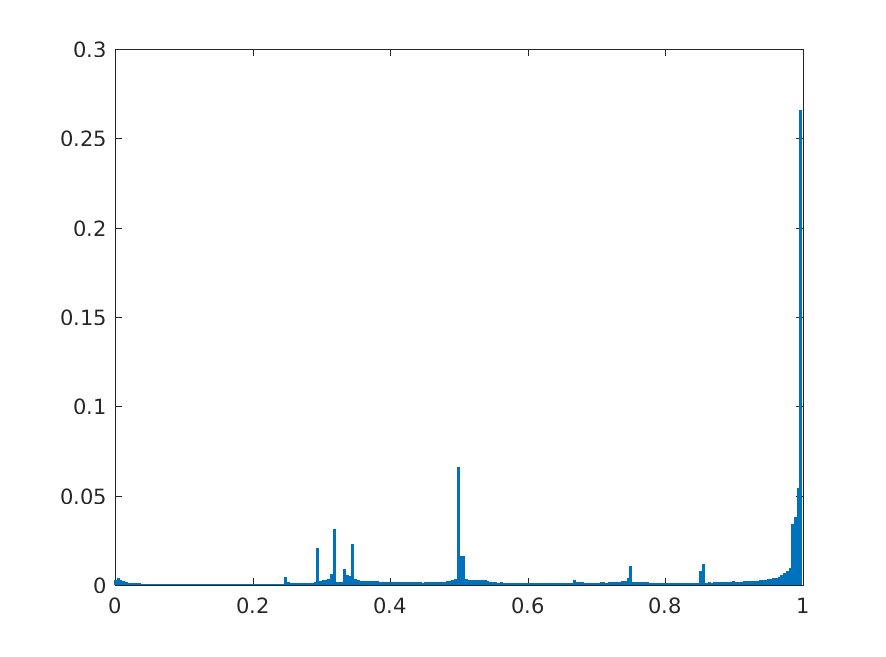}
\includegraphics[width=3.5cm, height=3cm]{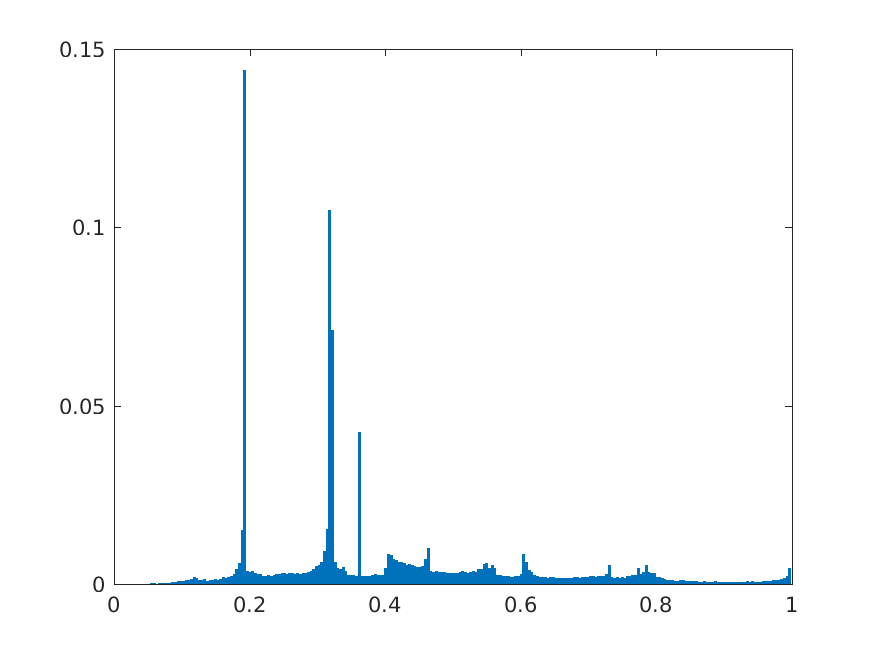}\\
(e)\hspace{3.5cm}(f)\\
\caption{(a) (b) Localizations obtained in images with Pulmonary Edema. (c) (d) Localizations obtained in normal images. (e) Average histogram of heat maps of CXRs with Pulmonary Edema. (f) Average histogram of heat maps for Pulmonary Edema detection in normal CXRs.}\label{fig:more_puled_loc}
\end{figure}

In a similar fashion, additional localization results for Pulmonary Edema is shown in Fig. \ref{fig:more_puled_loc}. In Fig. \ref{fig:more_puled_loc}(a) and (b) localization of two examples of CXRs with Pulmonary Edema is shown. As stated earlier the classifier localizes in the lung region. This is not the case when normal images are used to localize Pulmonary Edema as seen in Fig. \ref{fig:more_puled_loc}(c) and (d). The localizations are obtained in random dense locations such as the sternum or heart. Like the cardiomegaly case, the histogram averages for CXRs with pulmonary Edema (Fig. \ref{fig:more_puled_loc}(e)) shows a sensitivity toward pulmonary edema detection while the normal CXRs shows a spread out detection. It is interesting to note that, in the histogram of normal images high probability (>0.85) is non-existent, thus ensuring low false positive rate. In the test set none of the normal images have been diagnosed as Pulmonary Edema. The failure cases are shown in Fig. \ref{fig:card_loc_failures}. These CXRs are with Pulmonary Edema. However, the localization algorithm shows that one of them localizes in lungs whereas the other one shows a localization pattern similar to that obtained in normal CXRs.

\begin{figure}[h]
\centering
\includegraphics[width=3.5cm, height=3.5cm]{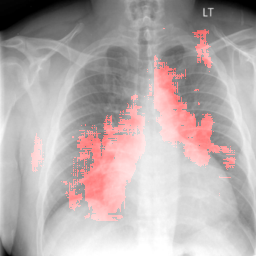}
\includegraphics[width=3.5cm, height=3.5cm]{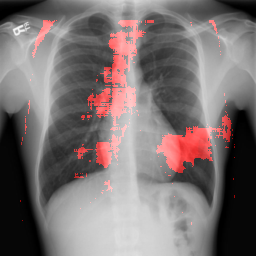}\\
(a)\hspace{3.5cm}(b)\\
\caption{Some examples of failure cases of Pulmonary Edema classification. (a) (b) Localization on CXRs with Pulmonary Edema which the classifier detects as normal.}\label{fig:puled_loc_failures}
\end{figure}

\end{document}